\documentclass{llncs}

\usepackage{mathpazo} 
\usepackage{eulervm}

\usepackage[utf8]{inputenc}
\usepackage[english]{babel}

\usepackage{amsmath}
\usepackage{amsthm}

\usepackage{latexsym,amsmath,amsfonts,amssymb,braket}
\usepackage{array}
\usepackage{listings}
\usepackage{ifmtarg}
\usepackage{calc}

\usepackage{amsfonts}
\usepackage{amssymb}
\usepackage{graphicx}
\usepackage{caption}
\usepackage{subcaption}
\graphicspath{{../images/}}
\usepackage{lmodern}
\usepackage{url}
\usepackage[noend]{algpseudocode}
\usepackage[table]{xcolor}
\usepackage[labelfont=bf]{caption}
\usepackage{listings}
\usepackage{courier}
\usepackage{enumitem}
\usepackage{emptypage}

\usepackage{siunitx}
\usepackage[table]{xcolor}
\usepackage{booktabs}
\usepackage{multirow}
\usepackage[ruled,vlined,linesnumbered]{algorithm2e}
\SetAlgoProcName{Class}{anautorefname}
\DontPrintSemicolon
\definecolor{vlgray}{gray}{0.92}

\usepackage{hyperref}

\usepackage{listings}
\usepackage{float}

\newfloat{pseudo}{tbph}{code}[section]
\floatname{pseudo}{Pseudocode}

\newfloat{code}{tbph}{code}[section]
\floatname{code}{Code}

\usepackage[scaled=0.80]{beramono}
\usepackage[T1]{fontenc}

\newlength{\MaxSizeOfLineNumbers}
\renewcommand*\thelstnumber{\hbox to 0em{\the\value{lstnumber}\hfil}}

\lstdefinestyle{lineNumber}{
	numbers=left, 
	numberstyle={\tiny}, 
	stepnumber=1, 
	numbersep=2em,
	numberblanklines=false
}

\makeatletter
\lst@Key{countblanklines}{true}[t]%
    {\lstKV@SetIf{#1}\lst@ifcountblanklines}

\lst@AddToHook{OnEmptyLine}{%
    \lst@ifnumberblanklines\else%
       \lst@ifcountblanklines\else%
         \advance\c@lstnumber-\@ne\relax%
       \fi%
    \fi}
\makeatother

\lstdefinelanguage{pseudo}{
	xleftmargin=3.6ex,
	aboveskip=1em,
	belowskip=1em,
	lineskip=1pt,
	mathescape,
	tabsize=2,
	basicstyle={\small},
	columns=fixed,
	morekeywords={
		abstract,case,catch,class,def,
		do,else,extends,false,final,finally,
		for,forall,all,if,implicit,import,match,mix,in,out,
		new,null,object,override,package,
		private,protected,requires,return,sealed,
		super,this,throw,trait,true,try,
		type,val,var,while,with,yield,to,until,
		then,and,or,not,function,method,procedure},
	otherkeywords={=},
	keywordstyle={\ttfamily \textbf},
	sensitive=true,
	morecomment=[l]{//},
	countblanklines=false,
	morestring=[b]"
}

\lstdefinelanguage{pseudoText}{
	xleftmargin=3.6ex,
	aboveskip=1em,
	belowskip=1em,
	lineskip=1pt,
	mathescape,
	tabsize=2,
	basicstyle={\small \ttfamily},
	columns=flexible,
	otherkeywords={=},
	keywordstyle={\ttfamily \textbf},
	sensitive=true,
	morecomment=[l]{//},
	morestring=[b]"
}

\lstdefinelanguage{scala}{
	xleftmargin=1.5em,
	aboveskip=1em,
	belowskip=1em,
	lineskip=1pt,
	tabsize=2,
	showstringspaces=false,
	basicstyle={\small \ttfamily},
	columns=fixed,
	morekeywords={
		abstract,case,catch,class,def,
		do,else,extends,false,final,finally,
		for,if,implicit,import,match,mix,in,
		new,null,object,override,package,
		private,protected,requires,return,sealed,
		super,this,throw,trait,true,try,
		type,val,var,while,with,yield,to,until,then,
		subjectTo,exploration,run,runSubjectTo},
	otherkeywords={!},
	keywordstyle={\ttfamily \textbf},
	sensitive=true,
	morecomment=[l]{//},
	commentstyle=\color{gray},
	countblanklines=false,
	morestring=[b]"	
}

\hypersetup{colorlinks=true, allcolors=blue}

\newcommand{\fig}[1]{Figure~\ref{fig:#1}}

\title{Kiwi --- A Minimalist CP Solver}

\author{
Renaud Hartert
}

\institute{UCLouvain, Belgium}

\date{}

\begin{document}

\thispagestyle{empty}

\maketitle
\vspace{-0,5cm}

\begin{abstract}
Kiwi is a minimalist and extendable Constraint Programming (CP) solver specifically designed for education. The particularities of Kiwi stand in its generic trailing state restoration mechanism and its modulable use of variables. By developing Kiwi, the author does not aim to provide an alternative to full featured constraint solvers but rather to provide readers with a basic architecture that will (hopefully) help them to understand the core mechanisms hidden under the hood of constraint solvers, to develop their own extended constraint solver, or to test innovative ideas.
\end{abstract}

\section{Introduction}

Nowadays, a growing number of real world problems are successfully tackled using constraint solvers. 
The hybridization of constraint solvers with other combinatorial technologies such as mixed integer programming~\cite{beck2003hybrid,bockmayr2003detecting,salvagnin2012hybrid}, local search~\cite{perron2004propagation,schaus13,lns1}, and particularly conflict driven clause learning~\cite{feydy2013semantic,feydy2009lazy,schutt2009cumulative} has been at the center of substantial advances in many domains.

Unfortunately, while many open-source constraint solvers exist~\cite{choco3,gecode,cpo,or-tools,oscar}, modifying these solvers to hybridize them with other technologies, to extend them with specific structured domains, or to enhance them with new functionalities, may prove to be a time consuming and discouraging adventure due to the impressive number of lines of code involved in those open-source projects. Also starting the development of a constraint solver from scratch may also prove itself to be quite a challenge even if one has a deep understanding of the mechanisms and techniques at play in constraint solvers and constraint propagators. 

The goal of this paper is to present \texttt{Kiwi} --- a minimalist and extendable constraint solver.
The source code of the core components of \texttt{Kiwi} is under 200 lines of Scala code and is the result of rethinking and simplifying the architecture of the open-source OscaR solver~\cite{oscar} to propose what we believe to be a good trade-off between performance, clarity, and conciseness.

By developing \texttt{Kiwi}, the author does not aim to provide an alternative to full featured constraint solvers 
but rather to provide students with a basic architecture that will (hopefully) help them to understand the core mechanisms hidden under the hood of constraint solvers, to develop their own extended constraint solver, or to test innovative ideas.\footnote{In this regard, \texttt{Kiwi} can be seen as an attempt to follow the initiative of \texttt{minisat}~\cite{minisat} but in the context of constraint programming.}

\section{Overview of the Solver}

We start the presentation of \texttt{Kiwi} by briefly describing its three main components: propagation, search, and state restoration.

\begin{paragraph}{Propagation}
The propagation system reduces the domain of the variables by filtering values that are part of no solution according to the constraints.
The architecture of this system is described in section~\ref{sec:kiwipropagation}. 
\end{paragraph}

\begin{paragraph}{Search} 
Unfortunately, propagation alone is usually not sufficient to solve a constraint satisfaction problem. 
Constraint solvers thus rely on a divide-and-conquer procedure that implicitly develops a search tree in which each node is a subproblem of its ancestors.
Leaves of this search-tree are either failed nodes -- i.e. inconsistent subproblems -- or solutions.
Propagation is used at the beginning of each node to reduce the domain of the variables and thus to prune fruitless branches of the search tree. 
The search procedure of \texttt{Kiwi} is described in section~\ref{sec:kiwisearch}.
\end{paragraph}

\begin{paragraph}{State restoration} 
The state restoration system manages the different search nodes explored during search. 
Precisely, it is the component that restores the domain of the variables to their previous state each time a backtrack occurs. 
Its main purpose is to reduce the cost of copying the entire state of each node to provide users with an efficient trade-off between memory and processing costs. 
State restoration mechanisms are presented in the next section.
\\
\end{paragraph}

\noindent
More than a constraint solver, \texttt{Kiwi} must be seen as a set of the components involved in the core of classic constraint solvers. 
Indeed, \texttt{Kiwi} does not provide users with abstraction to help them model and solve combinatorial problems.
Also, we only give little attention to the filtering procedures involved in constraint propagation  -- which are arguably the most complex parts of every constraint solver.
While the simplest binary constraints may only require a few lines of code, more complex global constraints usually rely on sophisticated algorithms and data structures. 
Scheduling constraints are a perfect example of such complex global constraints. 
The reader that is interested by global constraints and propagators may refer to~\cite{beldiceanu2007global} for a wide catalogue of global constraints and links towards the relevant literature. 

\section{State restoration}

Copying and storing the whole state of each search node is usually too memory consuming for a real constraint solver. 
The \emph{state restoration} system thus has to rely on different trade-offs to restore the domain of the variables as efficiently as possible.\footnote{State restoration is of course not limited to domains and may also be used to maintain incremental data structures or other components.}
There are three main approaches to implements such a system: 
\begin{itemize}
  \item \textbf{Copying.} A complete copy of the domains is done and stored before changing their state.
  \item \textbf{Recomputation.} Domains are recomputed from scratch when required.
  \item \textbf{Trailing.} Domain changes are recorded incrementally and undone when required.
\end{itemize}

\noindent
Trailing is the prominent approach used in many constraint solvers~\cite{choco3,minisat,cpo,or-tools,oscar}.\footnote{Although, some solvers such as Gecode~\cite{gecode} rely on an hybrid state restoration mechanism based on both copying and recomputation.}
The idea behind \emph{trailing} is to maintain the sequence of changes that occured on a given branch since the root of the search tree.
We call such a sequence the \emph{trail}.
Each node can be represented by a prefix of the trail that corresponds to the sequence of changes that lead to this node. 
Particularly, the sequence of changes that lead to a node is always an extension of the sequence that lead to its direct ancestor and so on (the root node being represented by the empty sequence). 
This incremental representation is convenient for backtracks as they can be performed by undoing the last changes of the trail until restoring the sequence of the desired node.

\begin{figure}
    \centering
    \includegraphics{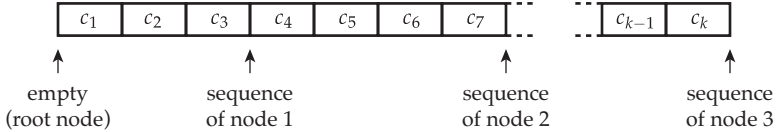}
    \caption{A trail made of a sequence of $k$ changes. The sequence of changes that lead to node $i$ is always an extension of the sequence of changes that lead to node $i-1$. }
    \label{fig:trailSeq}
\end{figure}

For instance, let us consider the sequence of $k$ changes depicted in \fig{trailSeq}. 
The first node has been computed from the root node by applying 3 changes $c_1, c_2, c_3$,
the second node has been computed from the first node by applying 4 changes $c_4, c_5, c_6, c_7$, 
and the third node has been computed from the second node by applying $k-7$ changes. 
We can easily restore the state of the second node by undoing the last changes of the trail until $c_7$, i.e., $c_k, c_{k-1}, \dots, c_8$.

\noindent
In \texttt{Kiwi}, we chose to manage state restoration with an extended trailing mechanism. 
The particularity of our system is that it allows each \emph{stateful object} (e.g. the domain of a variable) to manage its own restoration mechanism internally. 
This mechanism can be based on trailing, copying, or even recomputation.
The remainder of this section is dedicated to the implementation of this system.

\subsection{Changes as an Abstraction}

Each time a state changing operation is performed, the necessary information to undo this operation is stored on the trail. 
Such undo information can have many forms but is usually represented by a pair made of a memory location and of its corresponding value. 

In \texttt{Kiwi}, we chose to directly store the functions responsible of undoing state changes as first class objects, i.e., as closures.
Each stateful object thus relies on its own restoration mechanism which is handled by the closure itself.

The abstract class of an \emph{undo operation}, namely \texttt{Change}, is presented in Code~\ref{code:change}. It contains a single function \texttt{undo} which, as its name suggests, is responsible for undoing the change. 
\begin{code}
\begin{lstlisting}[language=scala, style=lineNumber]
abstract class Change {
  def undo(): Unit // Undo the change
}
\end{lstlisting}
\caption{The \texttt{Change} abstract class.}
\label{code:change}
\end{code}

\noindent
The complete implementation of a stateful mutable integer is presented in section~\ref{sec:trailedint}.

\subsection{The Trail}

Our trailing system is implemented with two stacks:
\begin{itemize}
  \item The first is a stack of \texttt{Change} objects that represents the trail.
  It is sorted chronologically such that the most recent change is on top of the stack (the root node being the empty stack).
  \item The second stack maps each node of the current branch to its corresponding prefix in the trail. 
  Thanks to the incrementality of the trail, only the position of the last change that lead to a node needs to be stored to characterize the whole sequence of changes that lead to this node. 
\end{itemize}

\begin{figure}
    \centering
    \includegraphics{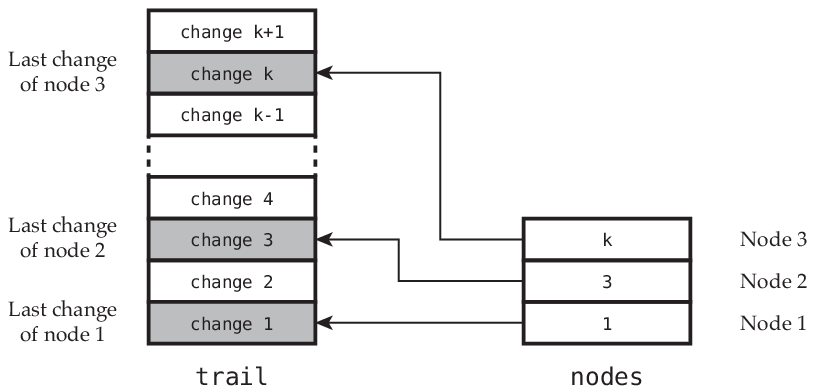}
    \caption{The trailing system of \texttt{Kiwi} is implemented with two stacks.}
    \label{fig:trailStacks}
\end{figure}

\noindent
\fig{trailStacks} illustrates the relation between both stacks. 

The whole implementation of \texttt{Trail}, our trailing system, is presented in Code~\ref{code:trailImplem}. 
Changes are registered using the \texttt{store} function that pushes them on top of \texttt{trail} (line 6).
The \texttt{newNode} function registers the state of the current node by storing the current size of the trail (line 8). 
Conversely, the \texttt{undoNode} function restores the previous node by undoing the changes on top of the \texttt{trail} until its size corresponds to the size stored with the previous node (lines 10 to 12 and 15 to 17).

\begin{code}
\begin{lstlisting}[language=scala, style=lineNumber]
class Trail {

  // Stacks to map nodes to the trail
  private val trail = new Stack[Change]()
  private val nodes = new Stack[Int]()
  
  // Store the change
  def store(change: Change): Unit = trail.push(change)
  
  // Mark the beginning of a new node
  def newNode(): Unit = nodes.push(trail.size)
  
  // Restore the previous node
  def undoNode(): Unit = {
    if (nodes.size > 0) undoUntil(nodes.pop())
  }
  
  // Restore the root node
  def undoAll(): Unit = undoUntil(0)
  
  private def undoUntil(size: Int): Unit = {
    while (trail.size > size) trail.pop().undo()
  }
}
\end{lstlisting}
\caption{Implementation of the trailing system of \texttt{Kiwi}}
\label{code:trailImplem}
\end{code}

\subsection{Trailed Integer}
\label{sec:trailedint}

We now have all the pieces required to build our first stateful object: a stateful integer variable\footnote{Integer variable here refers to a mutable integer.} called \texttt{TrailedInt}. 
As for classic integer variables, the value of \texttt{TrailedInt} can be accessed and updated. 
It however keeps track of the different values it was assigned to in order to restore them each time a backtrack occurs.

Similarly to most stateful objects in \texttt{Kiwi}, \texttt{TrailedInt} implements its own internal state restoration mechanism. 
It is based on a stack of integers that represents the sequence of values that were assigned to this \texttt{TrailedInt} since its initialization.
Restoring the previous state of \texttt{TrailedInt} thus amounts to update its current value to the last entry of the stack (which is then removed from the stack). 

The implementation of \texttt{TrailedInt} is presented in Code~\ref{code:trailedInt}.
The current value of the object is stored in the private variable \texttt{currentValue}. 
It is accessed with the \texttt{getValue} function. 
The \texttt{setValue} function is the one used to modify the value of the object. 
It pushes the current value on top of the stack of old values (line 7), updates \texttt{currentValue} (line~8), and notifies \texttt{Trail} that a change occurred (line 9). 

The particularity of this implementation is that \texttt{TrailedInt} directly implements the \texttt{Change} abstract class and its \texttt{undo} operation (lines 12 to 14). 
This has the advantage of reducing the overhead of instantiating a new \texttt{Change} object each time a state changing operation is performed on \texttt{TrailedInt}.

\begin{code}
\begin{lstlisting}[language=scala, style=lineNumber]
class TrailedInt(trail: Trail, initValue: Int) extends Change { 

  private val oldValues = new Stack[Int]()
  private var currentValue: Int = initialValue 
  
  def getValue: Int = currentValue
  
  def setValue(value: Int): Unit = {
    if (value != currentValue) {
      oldValues.push(currentValue)
      currentValue = value
      trail.store(this)
    }
  }
  
  override def undo(): Unit = { 
    currentValue = oldValues.pop() 
  }
}
\end{lstlisting}
\caption{Implementation of \texttt{TrailedInt}, a stateful integer variable with its own internal state restoration mechanism.}
\label{code:trailedInt}
\end{code}

\subsection{Improvements}

The trailing system we presented suffers from a major weakness.
For instance, \texttt{TrailedInt} keeps track of all the values it was assigned to. 
However, only the values it was assigned to at the beginning of each state are required for state restoration.
Hopefuly, we can easily fix this problem with \emph{timestamps}.
The idea is to associate a timestamp to each search node and to only register the current value of \texttt{TrailedInt} if it has not yet been registered at a given timestamp.
The interested reader can refer to the 12th chapter of~\cite{handbook} for detailed explanations on how to improve trailing systems. 

\section{The Propagation System}
\label{sec:kiwipropagation}

Propagation is the process of using the filtering procedure embedded in the constraints to reduce the domain of the variables. 
To achieve this, the propagation system of \texttt{Kiwi} relies on the following entities: 
\begin{itemize}

  \item The \emph{propagators} which are responsible of performing constraint propagation by updating the domain of the variables, or to notify that a conflict occured. They contain the actual filtering procedures of their respective constraint. Propagators are registered on different types of events, such as the reduction of a variable's domain, and need to be propagated each time one of these events occurs.  
  \item The \emph{propagation queue} that synchronizes propagators to perform propagation until no further filtering can be achieved or a conflict has been detected.

  \item The \emph{variables} which are updated by the propagators and are responsible for notifying the propagation queue if a propagator needs to be considered for further propagation. 
\end{itemize}

\begin{figure}
    \centering
    \includegraphics{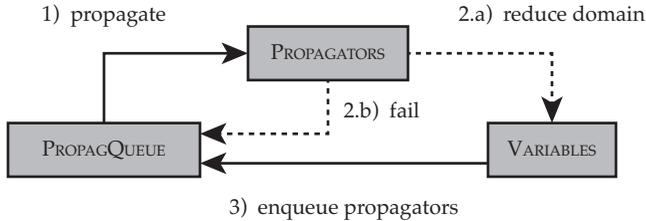}
    \caption{Interaction between the components at play in the propagation system of \texttt{Kiwi}.}
    \label{fig:propagSys}
\end{figure}

\noindent
The interaction between those abstractions is illustrated in~\fig{propagSys}. Observe that the propagation queue, \texttt{PropagQueue}, does not directly interact with the variables. 

\subsection{Propagators}

\texttt{Kiwi} does not contain any object that actually represents a constraint. 
Instead, a constraint is implicitly defined by the set of propagators which must ensure that the constraint's logical relation holds. 
A constraint can thus be composed of a single propagator registered on all the variables in the constraint's scope, or by a set of propagators, each one being registered on a subset of the constraint's scope.

The abstract class \texttt{Propagator} is presented in Code~\ref{code:propagator}. It contains the following functions:
\begin{itemize}
  \item The \texttt{enqueued} boolean is a flag used by the propagation queue to indicate that the propagator is awake and waiting for propagation. 

  \item The \texttt{init} function registers the propagator on the events of interest and perform its initial propagation. It return \texttt{true} if the propagation suceed and \texttt{false} if it lead to a conflict. 

  \item The \texttt{propagate} function performs the actual filtering. Like \texttt{init}, it returns \texttt{false} if and only if propagation lead to a conflict.
\end{itemize}

\begin{code}
\begin{lstlisting}[language=scala, style=lineNumber]
abstract class Propagator {
  var enqueued: Boolean    // True if the propagator is awake
  def init(): Boolean      // Register the propagator and propagate
  def propagate(): Boolean // Calls the filtering procedure 
}
\end{lstlisting}
\caption{The \texttt{Propagator} abstract class.}
\label{code:propagator}
\end{code}

\noindent
As an example, we present the complete implementation of a simple binary constraint in section~\ref{sec:leq}.

\subsection{The Propagation Queue}

The propagation queue, \texttt{PropagQueue}, contains all the propagators that are waiting for propagation. 
It is responsible for synchronizing these propagators until no further filtering can be achieved, or until a conflict has been detected. 
When running, the \emph{propagation process} dequeues one propagator from the propagation queue and calls its \texttt{propagate} function to reduce the domain of some variables. 
Of course, reducing the domain of a variable usually awakes new propagators that are then enqueued in the propagation queue. 
This process is repeated until either the propagation queue becomes empty, meaning that no further propagation is possible, or until a conflict occured.

The implementation of \texttt{PropagQueue} is presented in code \ref{code:solver}. 
The \texttt{enqueue} function enqueues a propagator in the propagation queue but only if it is not already contained in it (lines 5 to 10). 
The \texttt{propagate} function contains the main loop of the propagation process (lines 12 to 22). It empties the queue by dequeuing each propagator and calling their propagate function if no conflict has occurred yet. The function concludes by returning \texttt{true} if no conflict occurred; it returns \texttt{false} otherwise. Observe that the propagation queue is emptied even if a conflict occured. 

\begin{code}
\begin{lstlisting}[language=scala, style=lineNumber]
class PropagQueue {

  // The propagation queue
  private val queue = new Queue[Propagator]()
  
  // Enqueue the propagator if it is not already enqueued
  def enqueue(propagator: Propagator): Unit = {
    if (!propagator.enqueued) {
      propagator.enqueued = true
      queue.enqueue(propagator)
    }  
  }
  
  // Propagation process
  def propagate(): Boolean = {
    var unfailed = true
    // Propagate the propagators
    while (!queue.isEmpty) {
      val propagator = queue.dequeue()
      // Only call propagate if no conflict occurred
      unfailed = unfailed && propagator.propagate()
      propagator.enqueued = false
    }
    return unfailed
  }
}
\end{lstlisting}
\caption{Implementation of the propagation queue of \texttt{Kiwi}.}
\label{code:solver}
\end{code}

\subsection{Variables}

Variables are the last piece of our propagation system. 
Interestingly enough, \texttt{Kiwi} does not provide variables with an interface to implement. 
This design choice is one of the reasons that facilitates the extension of \texttt{Kiwi} with additional structured domain representations~\cite{handbook}.
While variables does not have to respect any requirements, they usually offer  the following functionalities:
\begin{itemize}
  \item A variable has a domain that represents the set of all possible values it could be assigned to. 
  \item A variable offers functions to remove possible values from its domain until it becomes a singleton, in which case the variable is assigned, or it becomes empty, meaning that a conflict occured. This domain must be restored to its previous state when a backtrack occurs. 
  \item A variable allows propagators to watch particular modifications of its domain. The role of the variable is then to enqueue these propagators in the propagation queue when one of these modifications occur.
\end{itemize} 

\noindent
As an example, we focus on a particular type of integer variable that we call \emph{interval variable}.\footnote{Interval variables are commonly used by \emph{bound-consistent} propagators~\cite{handbook}.}
The domain of an interval variable is characterized by its minimal value $min$ and its maximal value $max$.
It contains all the values contained in the interval $[min, max]$.
Particularly, the domain of an interval variable can only be reduced by increasing its minimal value or by decreasing its maximal value. 
We subsequently refer to the minimal and maximal values of the domain as the \emph{bounds} of the variables.

The implementation of \texttt{IntervalVar} is presented in Code~\ref{code:rangevar}.
As expected, it is made of two \texttt{TrailedInt} that represent the minimum and maximum value of the variable's domain.
The \texttt{minWatchers} and \texttt{maxWatchers} queues are used to store propagators that must respectively be awaken when the minimum value of the variable is increased or when the maximum value of the variable is decreased. 
These queues are filled using the \texttt{watchMin} and \texttt{watchMax} functions. 
The \texttt{updateMin} function is the one responsible for increasing the minimum value of the variable. 
It operates as follows:
\begin{enumerate}
  \item If the new minimum value exceeds the maximum value of the variable, then the domain becomes empty and the function returns \texttt{false} to notify this conflict (line 14).\footnote{We do not actually empty the domain of the variable because this change will be directly undone by backtracking.}
  \item If the new minimum value is lower or equal to the current minimum value, then nothing happens and the function returns \texttt{true} (line 15).
  \item Otherwise, the function updates the minimum value to its new value, awakes all the propagators contained in the \texttt{minWatchers} queue and returns \texttt{true} to notify that no conflict occured (lines 17 to 19). 
\end{enumerate}
For the sake of conciseness, we do not describe the implementation of \texttt{watchMax} and \texttt{updateMax} as they are symmetric to their \texttt{min} version.

\begin{code}
\begin{lstlisting}[language=scala, style=lineNumber]
class IntervalVar(pQueue: PropagQueue, trail: Trail, 
                  initMin: Int, initMax: Int) {

  private val min = new TrailedInt(trail, initMin)
  private val max = new TrailedInt(trail, initMax)
  private val minWatchers = new Queue[Propagator]()
  private val maxWatchers = new Queue[Propagator]()
  
  def getMin: Int = min.getValue
  def getMax: Int = max.getValue
  def isAssigned: Boolean = getMin == getMax
  
  def watchMin(propagator: Propagator): Unit = {
    minWatchers.enqueue(propagator)
  }
  
  def updateMin(newMin: Int): Boolean = {
    if (newMin > max.getValue) false
    else if (newMin <= min.getValue) true
    else {
      min.setValue(newMin)
      for (propagator <- minWatchers) pQueue.enqueue(propagator)
      true
    }
  }
  
  // Similar to watchMin
  def watchMax(propagator: Propagator): Unit = { ... }

  // Similar to updateMin
  def updateMax(newMax: Int): Boolean = { ... }
}
\end{lstlisting}
\caption{Implementation of an interval variable.}
\label{code:rangevar}
\end{code}

\subsection{The Lower or Equal Constraint}
\label{sec:leq}

We now have all the components at hand to understand and implement our first constraint.
We focus on the simple binary constraint $x \leq y$ where both $x$ and $y$ are interval variables (see Code~\ref{code:rangevar}).

The $x \leq y$ constraint is made of a single propagator called \texttt{LowerEqual} (see Code~\ref{code:leq}). Its \texttt{propagate} function ensures the following rules: 
\begin{itemize}
  \item The maximum value of $x$ is always lower or equal to the maximum value of $y$ (line 12).
  \item The minimum value of $y$ is always greater or equal to the minimum value of $x$ (line 13).
\end{itemize}
The \texttt{propagate} function returns \texttt{false} if ensuring these rules empties the domain of either $x$ or $y$; it returns \texttt{true} otherwise. 
The \texttt{init} function performs the initial propagation (line 4) and registers the propagator on its variables (lines 6 and 7). 
Like \texttt{propagate}, it returns \texttt{true} if the propagation succeeded and \texttt{false} if a conflict occurred. 
Note that the propagator does not have to be awaken when the maximum value of $x$ or the minimum value of $y$ change since they have no impact on both previous rules. 

\begin{code}
\begin{lstlisting}[language=scala,style=lineNumber]
class LowerEqual(x: IntervalVar, y: IntervalVar) extends Propagator {
  
  override def init(): Boolean = {
    if (!propagate()) false
    else {
      x.watchMin(this)
      y.watchMax(this)
      true
    }
  }

  override def propagate(): Boolean = {
    if (!x.updateMax(y.getMax)) false
    else if (!y.updateMin(x.getMin)) false
    else true
  }
}
\end{lstlisting}
\caption{Implementation of a propagator to enforce the relation $x \leq y$.}
\label{code:leq}
\end{code}

\subsection{Improvement and Discussion}

There are many ways to improve our propagation system. 
For instance, let us consider the \texttt{LowerEqual} propagator from the previous section. 
If the maximal value of $x$ becomes lower or equal to the minimum value of $y$, then no further filtering is possible.
We say that the propagator is \emph{entailed} and thus does not have to be propagated anymore. 
We could implement this additional condition by adding a \emph{stateful boolean} to the \texttt{Propagator} interface and by checking the value of this boolean each time the propagator is enqueued in the propagation queue. 
Another improvement is to add different levels of priority to propagators.
Indeed, some propagators have a much faster filtering procedure than others. 
It is therefore useful to propagate these propagators first in order to give as much information as possible to slower propagators. 
The interested reader can refer to the 4th and 12th chapters of~\cite{handbook} for detailed additional improvements. 

\section{Search}
\label{sec:kiwisearch}

The final part of our solver is the search system. 
Its aim is to explore the whole search space of the problem looking for solutions or proving that no solution exists. 
It uses trailing to manage the nodes of the search tree and applies propagation to prune fruitless branches. 
The search system is based on a depth-first-search algorithm that relies on \emph{heuristics} to determine the order in which nodes are explored.
The following sections are dedicated to the description of these components. 

\subsection{Heuristics and Decisions}

The aim of the heuristic is to define and order the children of any internal node of the search tree. 
It does so by associating each node with a sequence of \emph{decisions}, where a decision is any operation that transforms a node into one of its children -- e.g. assigning a variable to a value, or removing that value from the domain of the variable.
The order in which decisions are returned corresponds to the order in which children have to be visited in the search tree.
Particularly, the heuristic returns an empty sequence of decisions if the current node is a valid leaf, i.e., a solution.

Let us first take a look at the interface of a \texttt{Decision} (see Code~\ref{code:decision}). It contains the single function \texttt{apply} that is responsible of applying the decision. It returns \texttt{false} if applying the decisions directly lead to a conflict (e.g. if we try to assign a variable to a value that is not part of its domain for instance); it returns \texttt{true} otherwise. 

\begin{code}
\begin{lstlisting}[language=scala, style=lineNumber]
abstract class Decision {
  def apply(): Boolean
}
\end{lstlisting}
\caption{The \texttt{Decision} abstract class.}
\label{code:decision}
\end{code}

\noindent
As an example, we consider the implementation of both decisions in Code~\ref{code:decision2}.
The \texttt{GreaterEqual} decision updates the minimal value of $x$ to $v$. It returns \texttt{true} if $v$ is lower or equal to the maximal value of $x$; it returns \texttt{false} otherwise. 
The \texttt{Assign} decision assign the variable $x$ to the value $v$. 
It returns \texttt{true} if $v$ is part of the domain of $x$; it returns \texttt{false} otherwise. 

\begin{code}
\begin{lstlisting}[language=scala, style=lineNumber]
class GreaterEqual(x: IntervalVar, v: Int) extends Decision {
  override def apply(): Boolean = x.updateMin(v)
}

class Assign(x: IntervalVar, v: Int) extends Decision {
  override def apply(): Boolean = x.updateMin(v) && x.updateMax(v) 
}
\end{lstlisting}
\caption{Implementation of two decisions.}
\label{code:decision2}
\end{code}

\noindent
Code~\ref{code:heuristic} presents the interface of \texttt{Heuristic}. 
It contains the single function \texttt{nextDecisions} which, as mentioned above, returns a sequence -- an iterator precisely -- of \texttt{Decision} objets that will be used by the search system to build the search tree.
\begin{code}
\begin{lstlisting}[language=scala, style=lineNumber]
abstract class Heuristic {
  def nextDecisions(): Iterator[Decision]
}
\end{lstlisting}
\caption{The \texttt{Heuristic} abstract class.}
\label{code:heuristic}
\end{code}

\begin{figure}
    \centering
    \includegraphics{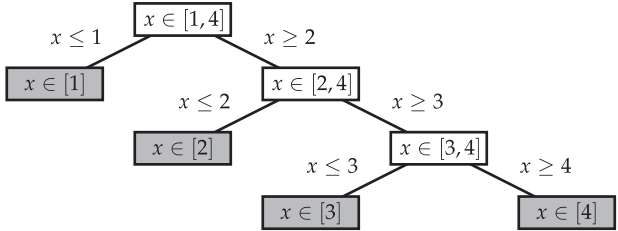}
    \caption{Search tree built by the \texttt{StaticMin} heuristic on variable $x \in [1, 4]$.}
    \label{fig:staticMin}
\end{figure}

\noindent
We illustrate the use of this interface with the \texttt{StaticMin} heuristic (see Code~\ref{code:staticMin}). 
It builds a binary search tree in which each left branch tries to assign a variable to its minimal value and each right branch tries to remove that value from the domain of its variable (see \fig{staticMin}). 
The order in which \texttt{StaticMin} tries to assign variables is statically determined by the order of the variables in the \texttt{vars} array. 
The first part of the heuristic (lines 6 to 7) searches for the index of the first unassigned variable in \texttt{vars}. 
This index is stored in the \texttt{nextUnassigned} stateful integer (line 8) -- the use of a stateful integer is not required here but reduces the overhead of systematically scanning the first already assigned variables. 
If all variables have been assigned, the heuristic returns an empty iterator to inform the search system that the current node is a leaf (line 10).
Otherwise, it selects the next unassigned variable and tries to assign it with its minimum value -- left branch -- or to remove that value from the variable's domain  -- right branch -- (lines 12 to 15).\footnote{The \texttt{LowerEqual} and the \texttt{GreaterEqual} decisions have similar implementations.}

\begin{code}
\begin{lstlisting}[language=scala, style=lineNumber]
class StaticMin(trail: Trail, vars: Array[IntervalVar]) 
                extends Heuristic {

  private val nextUnassigned = new TrailedInt(trail, 0)

  override def nextDecisions: Iterator[Decision] = {
    // Search for the next unassigned variable
    var i = nextUnassigned.getValue
    while (i < vars.length && vars(i).isAssigned) i += 1
    nextUnassigned.setValue(i)
    // Compute the sequence of decisions
    if (i == vars.length) Iterator() // leaf node
    else {
      val x = vars(i)
      val v = x.getMin
      // Assign x to v or remove v from the domain of x
      Iterator(new LowerEqual(x, v), new GreaterEqual(x, v + 1))
    }
  }
}
\end{lstlisting}
\caption{Implementation of a static binary search heuristic that tries to assign variables to their minimum value.}
\label{code:staticMin}
\end{code}

\subsection{The Depth-First-Search}
\label{sec:dfs}

Our \emph{depth-first-search} algorithm explores the search tree defined by the heuristic. 
It relies on propagation to reduce the domain of the variables and thus pruning fruitless branches of the search tree. 

\begin{figure}
    \centering
    \includegraphics{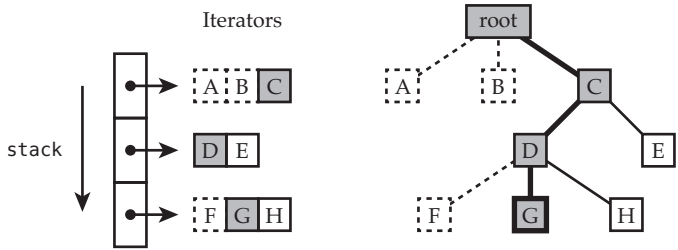}
    \caption{Relation between the stack of iterators and the search tree.}
    \label{fig:dfsStack}
\end{figure}

The particularity of our implementation is that it manages its own stack of nodes instead of relying on recursion. 
The link between the stack and the actual search tree is illustrated in \fig{dfsStack} (note that, by abuse of language, we subsequently refer to `decisions' as `nodes' and vice-versa).
The stack (depicted on the left) contains the iterator of decisions of each node in the currently explored branch of the search tree. 
Precisely, the first iterator corresponds to the children of the root node, the second iterator corresponds to the children of node $C$, and so on. 
The gray element of each iterator is its current element. Hence, dashed elements have already been read while white elements still have to be. 
The search tree (depicted on the right) follows the same color code. 
Dashed nodes are the root of already explored subtrees, gray nodes are part of the branch that is currently explored by the search, and white nodes still have to be visited. 
Node $G$ is the node that is currently visited by the search. 

\begin{figure}
    \centering
    \includegraphics{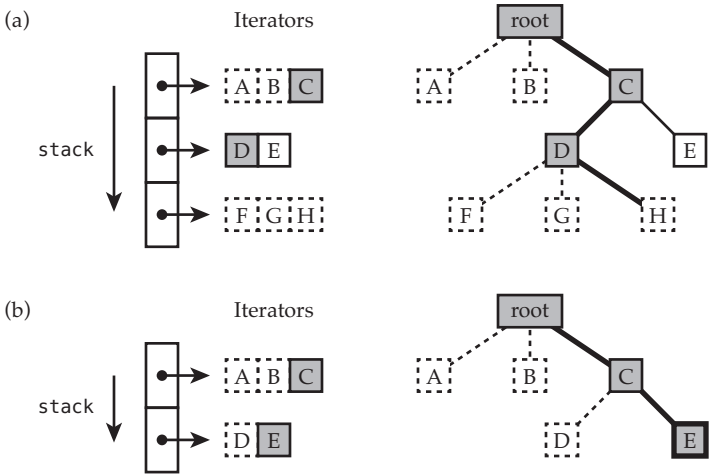}
    \caption{Relation between the stack and the search tree in case of backtrack.}
    \label{fig:dfsStack2}
\end{figure}

The stack starts empty at the beginning of the search to indicate that the root node has not yet been visited. 
The search then uses the heuristic to \emph{expand} the root node by pushing its sequence of children on the stack. 
The first unvisited child on top of the stack is then selected and expanded in the same way. 
This process is repeated until, eventually, the search reaches a node with an empty sequence of decisions, i.e., a leaf. 
It then operates as follows:
\begin{enumerate}
	\item If the iterator on top of the stack still contains unvisited nodes, then the search backtracks by visiting the next element of this iterator --~the next sibling of the current node. 
	\item Otherwise, we know that the iterator on top of the stack does no contain any unvisited nodes. The search thus removes this iterator from the top of the stack and repeats this process.
\end{enumerate}
\fig{dfsStack2} illustrates this operation where steps (a) and (b) respectively correspond to the second and the first above situations. 
The search keeps exploring the search tree by pushing and popping iterators on the stack until it becomes empty, meaning that the whole search tree has been explored. 

The implementation of our depth-first-search algorithm is presented in Codes~\ref{code:dfs1} and~\ref{code:dfs2}.
As expected, it contains an internal stack of iterators that we call \texttt{decisions} (line 2). 
The \texttt{onSolution} function (line 4) is a hook to be overridden by the user. 
It defines the action to perform each time the search encounter a solution. 
The \texttt{propagateAndExpand} function (lines 5 to 15) is called each time a new node is visited. It returns \texttt{false} if and only if the currently visited node is a leaf (either it is a solution or not); it returns \texttt{true} otherwise.
The first part of that function is to propagate the visited node (line 6). 
This operation may lead to a conflict in which case \texttt{false} is returned to indicate that the node is a failed leaf. 
Otherwise, we call the \texttt{nextDecisions} function of \texttt{heuristic} to get the children of the node as a sequence of decisions (line 7). 
If this sequence is empty, then we know that the node is a solution, we call the \texttt{onSolution} function, and returns \texttt{false} (lines 8 to 10). 
If the node actually has children, its iterator of decisions is pushed on the stack and the function returns \texttt{true} (lines 11 to 13).

\begin{code}
\begin{lstlisting}[language=scala, style=lineNumber]
class Search(solver: Solver, trail: Trail, heuristic: Heuristic) {

  private val decisions = new Stack[Iterator[Decision]]()
  
  // Called each time a solution is found
  def onSolution(): Unit = {}
  
  private def propagateAndExpand(): Boolean = {
    if (!solver.propagate()) return false
    val nextDecisions = heuristic.getDecisions()
    if (nextDecisions.isEmpty) {
      onSolution()
      false
    else {
      decisions.push(nextDecisions)
      true
    }
  }
  
  def search(maxNodes: Int): Boolean = { ... }
}
\end{lstlisting}
\caption{Solver}
\label{code:dfs1}
\end{code}

\noindent
The \texttt{search} function is probably the most complex piece of code of \texttt{Kiwi} (see Code~\ref{code:dfs2}). 
It is the one that is actually responsible of exploring the search tree. 
It takes a maximum number of nodes to visit in parameter to limit the time spent exploring the search tree --~which grows exponentially with the size of the problem to solve. 
The function returns \texttt{true} if it was able to explore the whole search tree without exceeding the maximum number of nodes; it returns \texttt{false} otherwise. 
The first part of the search is to call \texttt{propagateAndExpand} to perform the initial root propagation and expand the root node (line 4). 
If that call returns \texttt{false}, then we know that the root node is already a leaf (either it is a solution or not).
The search thus returns \texttt{true} to notify the user that the whole search tree -- which is made of a single node -- has been explored.
Otherwise, we know that the \texttt{decisions} stack contains its first non-empty iterator (i.e. the children of the root) and that we can start exploring the search tree. 
The actual search starts if the \texttt{decisions} stack contains at least one iterator -- which is always true at this step -- and if the maximum number of nodes is not exceeded (line 6).
The first step of the loop is to check that the iterator on top of the stack still contains unapplied decisions, i.e., unvisited nodes (lines 7 and 8).
The next step is one of the following:
\begin{enumerate}

	\item If the iterator still contains some decisions, then the search visits the next node by notifying the trailing system that we are building a new node (line 10), and then by applying the next decision of the top iterator (lines 11 and 12). 
The search undo the state of this new node if it is a leaf (line~13), i.e., if applying the decision directly lead to a conflict or if the \texttt{propagateAndExpand} function returned \texttt{false}.

	\item If the iterator does not contain any unapplied decision, then it is removed from the \texttt{decisions} stack (line 15) and the state of the current node is undone (line 16) (see \fig{dfsStack2}). 
	
\end{enumerate}
The main loop is repeated until either the \texttt{decisions} stack becomes empty or until the maximum number of visited nodes has been exceeded. 
The search then concludes by undoing all the not yet undone nodes and by removing all the remaining iterators from \texttt{decisions} (lines 20 and 21). 
It finally returns \texttt{true} if and only if it was able to explore the whole search tree (line 23). 

\begin{code}
\begin{lstlisting}[language=scala, style=lineNumber] 
def search(maxNodes: Int): Boolean = {
  var nodes = 0
  // Root propagation
  if (!propagateAndExpand()) return true
  // Start DFS
  while (!decisions.isEmpty && nodes < maxNodes) {
    val nextDecisions = decisions.top
    if (nextDecisions.hasNext) {
      nodes += 1
      trail.newNode()
      val decision = nextDecisions.next()
      val success = decision.apply()
      if (!success || !propagateAndExpand()) trail.undoNode()
    } else {
      decisions.pop()
      trail.undoNode()
    }
  }
  // Clear trail and decisions
  trail.undoAll()
  decisions.clear()
  // Return true if the search is complete
  return nodes < maxNodes
}  
\end{lstlisting}
\caption{The depth-first-search algorithm of \texttt{Kiwi}.}
\label{code:dfs2}
\end{code}

\bibliographystyle{plain}
\bibliography{paper}

\end{document}